%% file: main.tex
\documentclass[11pt]{article}

\usepackage{style}
\usepackage{times}
\usepackage{url}
\usepackage[numbers]{natbib}
\usepackage{latexsym}
\usepackage{graphicx}
\usepackage{subfiles}
\usepackage[colorlinks=true,urlcolor=blue,linkcolor=blue,anchorcolor=blue,citecolor=blue]{hyperref}

\title{An Open-Source American Sign Language Fingerspell Recognition and Semantic Pose Retrieval Interface}

\author{Kevin Jose Thomas \\
  Burnaby South S. S \\
  Burnaby, BC, Canada\\
  {\tt kevin.jt2007@gmail.com} \\
}

\begin{document}
\maketitle
\begin{abstract}
  This paper introduces an open-source interface for American Sign Language fingerspell recognition and semantic pose retrieval, aimed to serve as a stepping stone towards more advanced sign language translation systems. Utilizing a combination of convolutional neural networks and pose estimation models, the interface provides two modular components: a recognition module for translating ASL fingerspelling into spoken English and a production module for converting spoken English into ASL pose sequences. The system is designed to be highly accessible, user-friendly, and capable of functioning in real-time under varying environmental conditions like backgrounds, lighting, skin tones, and hand sizes. We discuss the technical details of the model architecture, application in the wild, as well as potential future enhancements for real-world consumer applications.
\end{abstract}

\section*{Acknowledgements}
As a hearing ASL student with elementary proficiency, I recognize that my perspective as a hearing person is limited. My role has been to listen carefully and integrate feedback from the Deaf community, and I have done my best to approach this project with a mindset of learning and understanding. This project would not have been possible without the active involvement and advice from ASL experts who have generously shared their insights.

It is also important to note that fingerspelling is only one aspect of ASL\footnote{Fingerspelling is the method of sign language where words are individually spelled out using hand movements. Fingerspelling accounts for up to 35\% of expressed ASL \cite{fingerspelling_stats}}, and the system does not aim to replace the richness and complexity of ASL grammar and syntax. Instead, it is designed to be a stepping stone towards more advanced ASL translation systems.

\subfile{Introduction.tex}

\subfile{Background.tex}

\subfile{Method.tex}

\subfile{Recognition.tex}

\subfile{Production.tex}

\subfile{Interface.tex}

\subfile{FutureWork.tex}

\subfile{Conclusion.tex}

\bibliographystyle{plainnat}
\renewcommand{\bibfont}{\small}
\setlength{\bibsep}{0pt}

\bibliography{refs}

\end{document}

%% file: Introduction.tex
\section{Introduction}
American Sign Language (ASL) is a complete, natural language that exhibits the same linguistic complexities as spoken languages, including its own syntax, morphology, and grammar that significantly differ from English. ASL is the primary language of many individuals in North America, and it is used by millions of people worldwide \cite{asl_stats}. Despite its prevalence, there remains a significant lack of resources and tools available to the public to facilitate communication between ASL and spoken language users.

This paper presents an open-source American Sign Language fingerspell recognition and semantic pose retrieval system with two main components:

\begin{itemize}
    \item \textbf{Recognition:} Ability to interpret fingerspelling and express it as spoken English text
    \item \textbf{Production:} Ability to interpret spoken English and express it as signing pose\footnote{A pose is a skeletal representation of a person's body} sequences
\end{itemize}

To serve as a stepping stone towards more real-world sign language translation systems, the system is designed to be modular and adaptable. The two components can be used independently or together, and the system can be implemented into various existing applications through a WebSocket API. Similarly, the system is designed to be accessible in real-world scenarios, where lighting conditions, skin tones, hand sizes, and distances from the camera can vary significantly. All the code and models are open-source and available on GitHub\footnote{All code and models are available at \url{https://github.com/kevinjosethomas/sign-language-processing}}.

This documented approach utilizes a combination of existing technologies and custom models to achieve the desired functionality. The recognition component provides two alternative models to translate fingerspelling. The first model is a custom 2D convolutional classification model built for simplicity and speed. The second model is an implementation of the 3D PointNet \cite{PointNet} architecture, which is more complex but provides better accuracy. Classified letters are synthesized into spoken English text using rule-based algorithms, and syntactical misclassifications are corrected using a pretrained BERT model \cite{Neuspell}. The production component uses a large language model (LLM) to translate spoken English text to ASL gloss\footnote{ASL gloss is a system of representing ASL signs and non-manual markers in written form}. Then, it uses semantic search to find corresponding ASL poses for each gloss, which are then synthesized into a stitched ASL pose sequence.

Furthermore, we package the recognition and production components into a WebSocket API that can be individually accessed by any user interface. We also develop an open-source user-friendly interface that allows for users to interact with both the modules in real-time.

%% file: Background.tex
\section{Background}

Sign language processing (SLP) is a field of research that focuses on the development of technologies that can understand and generate sign language. With the rise of novel deep learning and computer vision technology, the field of SLP has seen significant advancements in recent years. Sign language processing can be segmented into two main categories: recognition and production. Sign language recognition focuses on the understanding of sign language, while sign language production focuses on the generation of sign language \cite{SLP}. In this section, we provide an overview of the current state-of-the-art in both receptive and expressive SLP.

\subsection{Recognition}

There are multiple components to the reception of sign language, including fingerspell recognition, sign language detection, sign language segmentation, and sign language recognition \cite{SLP}. Fingerspell recognition involves recognizing individual ASL letters in videos or images. Sign language detection involves recognizing whether a given video or image contains sign language. Sign language segmentation involves isolating different signs in a given video. Sign language recognition refers to the process of labeling signs in videos. While all tasks are essential to the overarching goal of sign language translation, it is important to distinguish between them.

\subsubsection*{Fingerspell Recognition}
\citet{SpellingItOut} developed an ASL fingerspelling recognition interface that utilizes a Microsoft Kinect device to capture real-time depth and appearance images. They found that classification using both depth and appearance images outperformed classification using only depth images. \citet{FingerspellingInTheWild} used videos from uncontrolled environments to train a fingerspell recognition model that utilized an iterative attention mechanism to improve performance. Furthermore, they utilized this mechanism to create two new data sets of annotated fingerspelling videos in the wild \cite{chicago2,chicago1}.

\subsubsection*{Sign Language Detection}
Sign language detection is the binary classification task of determining whether a given video features someone signing. With the goal of spotlighting signers when they sign in video conferencing software, \ \citet{SignLanguageDetection} utilized optical flow features from OpenPose human pose estimation to train a binary classifier. They produced results of 87\%-91\% accuracy with a per-frame inference time of only 350–3500\textit{\textmu}s. \citet{SignLanguageDetectionInTheWild} used a multi-layer recurrent neural network and a 2-stream convolutional neural network to analyze video and motion data for sign language detection. They achieved an accuracy of 87.67\% which outperformed the previous state-of-the-art baseline model by 18.44\%.

\subsubsection*{Sign Language Segmentation}
\citet{SignLanguageSegmentation} described a novel approach to sign language segmentation, focusing on segmenting individual signs and phrases. They introduced a method that incorporated optical flow features to better capture phrase boundaries. \citet{TemporalSignLanguageSegmentation} developed a 3D convolutional neural network to identify temporal boundaries between signs in continuous sign language videos.

\subsubsection*{Sign Language Translation}
\citet{PopSign} developed PopSign ASL, the first sign language translation system available to the public. It featured an LSTM model that could translate individual signs into text with an accuracy of 84.2\%. Furthermore, they created a data set of 250 isolated signs, comprising over 210,000 individual videos from 47 different signers.

\subsection{Production}
Similar to recognition, sign language production is often broken down to multiple sub-tasks to achieve the desired functionality. Current sign language production systems follow the pipeline of text-to-gloss-to-pose-to-video \cite{SLP}. 

\subsubsection*{Text to Gloss}
\citet{Text2Gloss2Pose2Video} explored three different methods of translating spoken English text to ASL gloss: a lemmatizer, a rule-based word reordering and dropping system, and a neural machine translation model. In their work, they opted for the rule-based word reordering system. \citet{TransformerText2Gloss} integrated syntactic information from a dependency parser into word embeddings to develop a syntax-aware transformer model to translate spoken English text to ASL gloss. They found that incorporating syntactic data resulted in improved translational accuracy and performance.

\subsubsection*{Gloss to Pose}
\citet{Gloss2Pose} built a lookup-table which mapped sign glosses to 2D skeletal poses, allowing them to play back sign poses from a given gloss. \citet{Text2Gloss2Pose2Video} used MediaPipe Holistic to extract body pose estimations from sign language datasets, and then stitched poses together to create poses from given glosses.

\subsubsection*{Pose to Video}
\citet{Text2Gloss2Pose2Video} used a Pix2Pix model and a ControlNet model to generate videos of people signing from given poses. They found that ControlNet with AnimateDiff outperformed the Pix2Pix model in terms of realism and quality of generated videos.

%% file: Method.tex
\section{Method}

In this section, we describe the technical pipelines of the recognition and production components of our interface. In-depth details on the data collection, model training, and inference stages for the two components are in \autoref{sec:recognition} and \autoref{sec:production}, respectively.

\subsection{Fingerspell Recognition}

Recognition refers to the process of interpreting sign language from a given video or image. In the context of this project, it involves recognizing fingerspelling and translating it into spoken English text. Traditionally, an image classification convolutional neural network (CNN) would be used to recognize fingerspelling. However, with the goal being real-time fingerspell reception in the real world, this component must be fast and accurate, regardless of variability in lighting conditions, backgrounds, skin tones, hand sizes, and distances from the camera. A model trained on images would not be invariant to these factors.

\begin{figure}[!htbp]
  \centerline{\includegraphics[width=\linewidth]{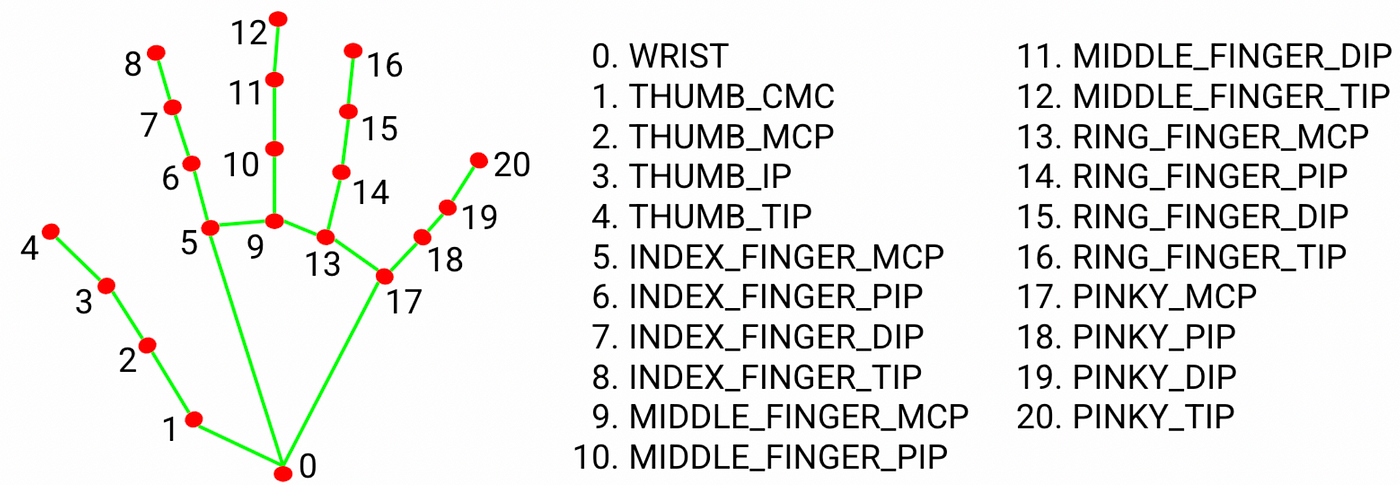}}
  \caption{Google MediaPipe hand landmarks}\label{fig:mediapipe-hand-landmarks}
\end{figure}

To combat this, we used the Google MediaPipe Hand Landmark \cite{MediaPipe} model, allowing for the real-time recognition and pose estimation of hand movements and positions. As seen in \autoref{fig:mediapipe-hand-landmarks}, MediaPipe captures 21 3D landmarks in given images, and it provides detailed information on handedness, hand orientation and finger position of hands in the frame. Furthermore, it can run locally on web browsers and does not require significant computational resources to run in a real-time setting. The use of MediaPipe also allows for the model to be invariant to varying lighting conditions, backgrounds, skin tones, hand sizes, and distances from the camera. By training a model on hand landmarks of fingerspelling images rather than the images themselves, we can achieve a model with low inference time that is invariant to these factors, suitable for real-time fingerspell reception in real-world scenarios.

\subsubsection*{Recognition Pipeline}

\begin{figure}[!htbp]
  \centerline{\includegraphics[width=\linewidth]{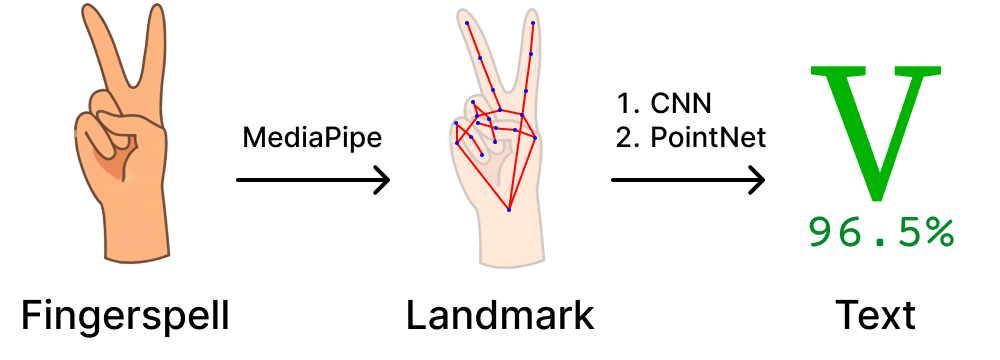}}
  \caption{Fingerspell recognition pipeline}\label{fig:recognition_method}
\end{figure}

The recognition component must take sequences of frames as input, classify the fingerspelling in each individual frame, and then synthesize the classified English alphabets into spoken English text. For each input frame $\mathbf{F_1, F_2, F_3,...,F_x}$, the model must output the corresponding English alphabet $\mathbf{A_1, A_2, A_3,...,A_y}$. Since a single fingerspelled alphabet can span multiple frames, we can assume $x \geq y$. This process is enabled by a three-step pipeline, as demonstrated in \autoref{fig:recognition_method}: fingerspelling-to-landmark conversion, landmark-to-text classification, and text synthesis.

\subsection{Production}

\begin{figure}[!htbp]
  \centerline{\includegraphics[width=\linewidth]{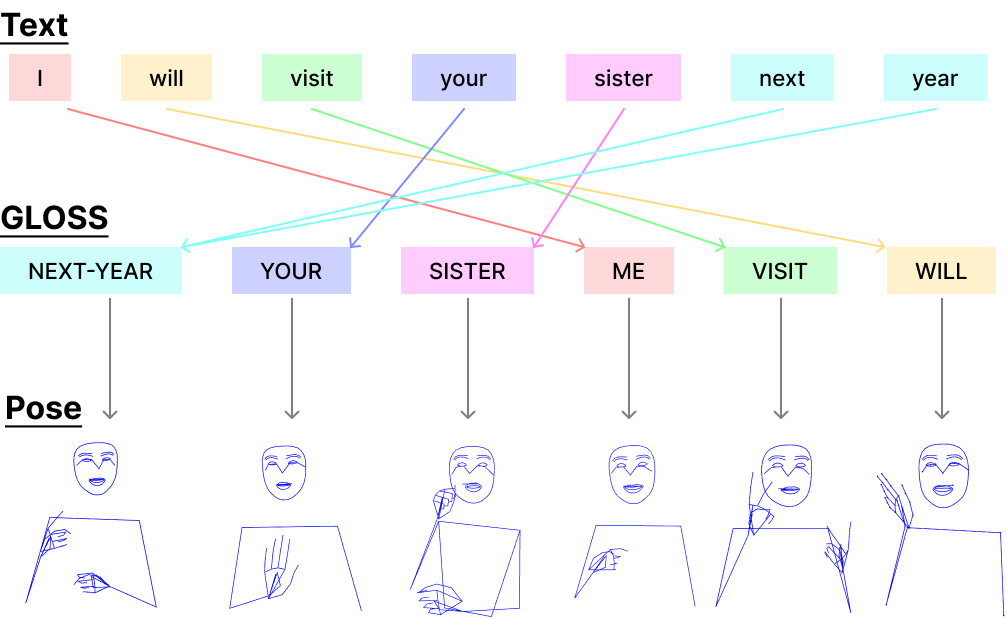}}
  \caption{Text \textrightarrow\ Gloss \textrightarrow\ Pose pipeline}\label{fig:production_method}
\end{figure}

Sign language production is the process of generating sign language videos from a given text. In the context of this project, production involves translating spoken English text to ASL gloss and generating sequences of animations of the corresponding ASL poses. Traditionally, we would use a simple lookup table to map glosses to poses. However, one of the main challenges in sign language production is the lack of large-scale data sets or sign language corpora. To combat this, we use semantic search to find contextually-similar poses for each gloss, even if the gloss is not present in the data set. This complete sign language production task is enabled by a two-step pipeline, as demonstrated in \autoref{fig:production_method}: text-to-gloss translation and semantic pose retrieval.

%% file: Recognition.tex
\section{Recognition}
\label{sec:recognition}

\subsection{Data Collection and Augmentation}

\subsubsection*{Source of Data}
The success of any machine learning project is heavily dependent on the quality and quantity of the data collected. To create a robust data set, we sourced approximately 250,000 images of ASL fingerspelling from three publicly accessible Kaggle data sets \cite{Kaggle1, Kaggle2, Kaggle3}. These images covered the full spectrum of ASL alphabets, ensuring that the model could accurately recognize each letter in different hand positions and varying distances from the camera. The diversity of these images was crucial to account for the variability in real-world scenarios. Additionally, we removed training data for the dynamic letters $\mathbf{\{J, Z\}}$ as their fingerspelling involve movement, which would require a separate model to recognize.

\begin{figure}[!htbp]
  \centerline{\includegraphics[width=\linewidth]{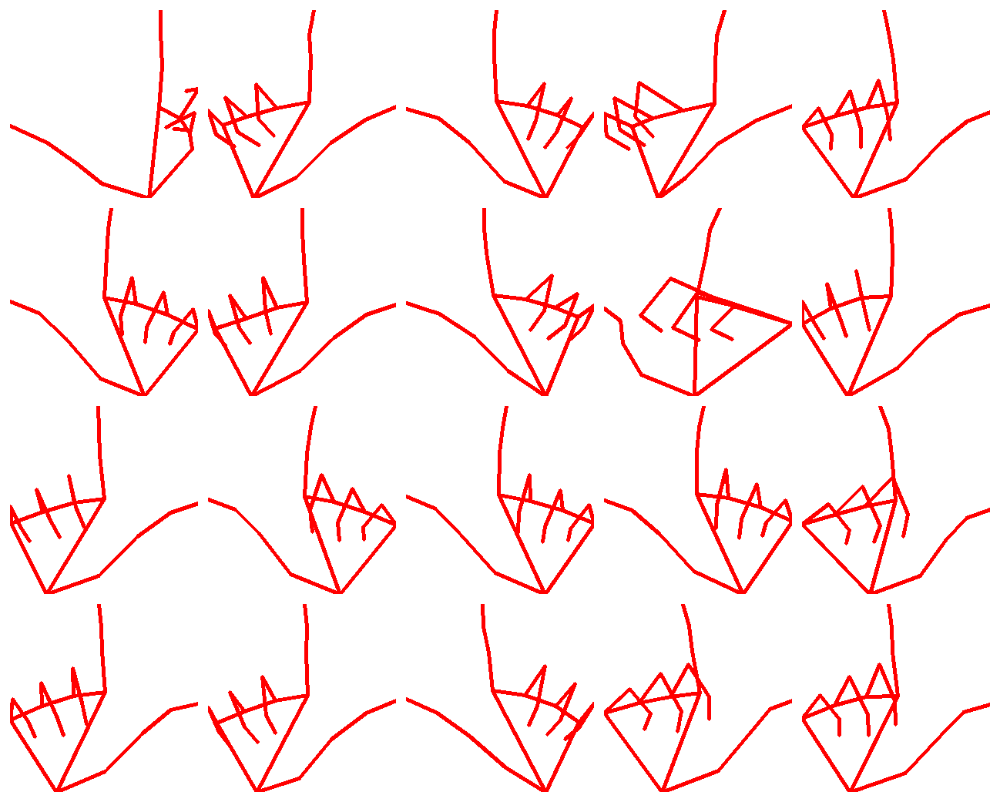}}
  \caption{Processed training data for the letter \lq L\rq}\label{fig:net_gallery}
\end{figure}

\subsubsection*{Preprocessing}
To ensure that the model maintained accuracy regardless of variability and to ensure it functioned in real-time, the data was converted to Google MediaPipe \cite{MediaPipe} hand landmarks and stored as NumPy \cite{numpy} arrays. This processing reduced the size of the data set to approximately 150,000 images where MediaPipe could accurately detect the 21 landmarks. Then, the detected landmarks were also scaled and normalized to a common reference frame, ensuring consistency across different images. Overall, the use of Google MediaPipe and the utilized normalization techniques made the data invariant to varying backgrounds, lighting conditions, skin tones, hand sizes, and distances from the camera. \autoref{fig:net_gallery} shows a visualized sample of the processed training data for the letter \lq L\rq.

\subsection{Model Training}

\begin{figure}[!htbp]
  \centerline{\includegraphics[width=\linewidth]{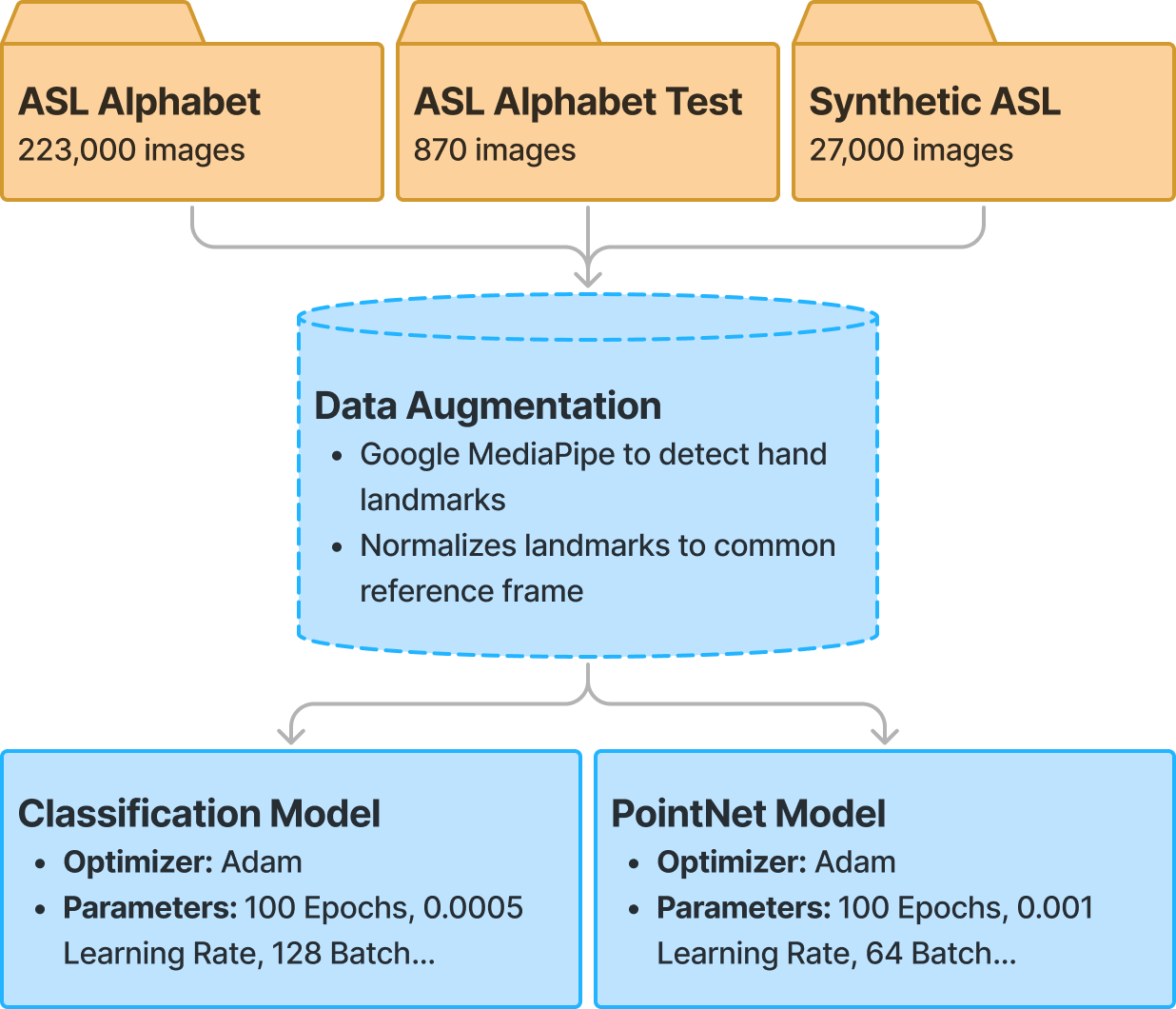}}
  \caption{Recognition training flow}\label{fig:recognition_training}
\end{figure}

As seen in \autoref{fig:recognition_training}, we use the augmented data set to train two classification models: a custom 2D convolutional classification model and an implementation of the 3D PointNet architecture \cite{PointNet}. The custom 2D convolutional classification model was built for simplicity and speed, while the 3D PointNet model was more complex but provided better accuracy.

\subsubsection*{2D CNN Architecture}

\begin{figure}[!htbp]
  \centerline{\includegraphics[width=\linewidth]{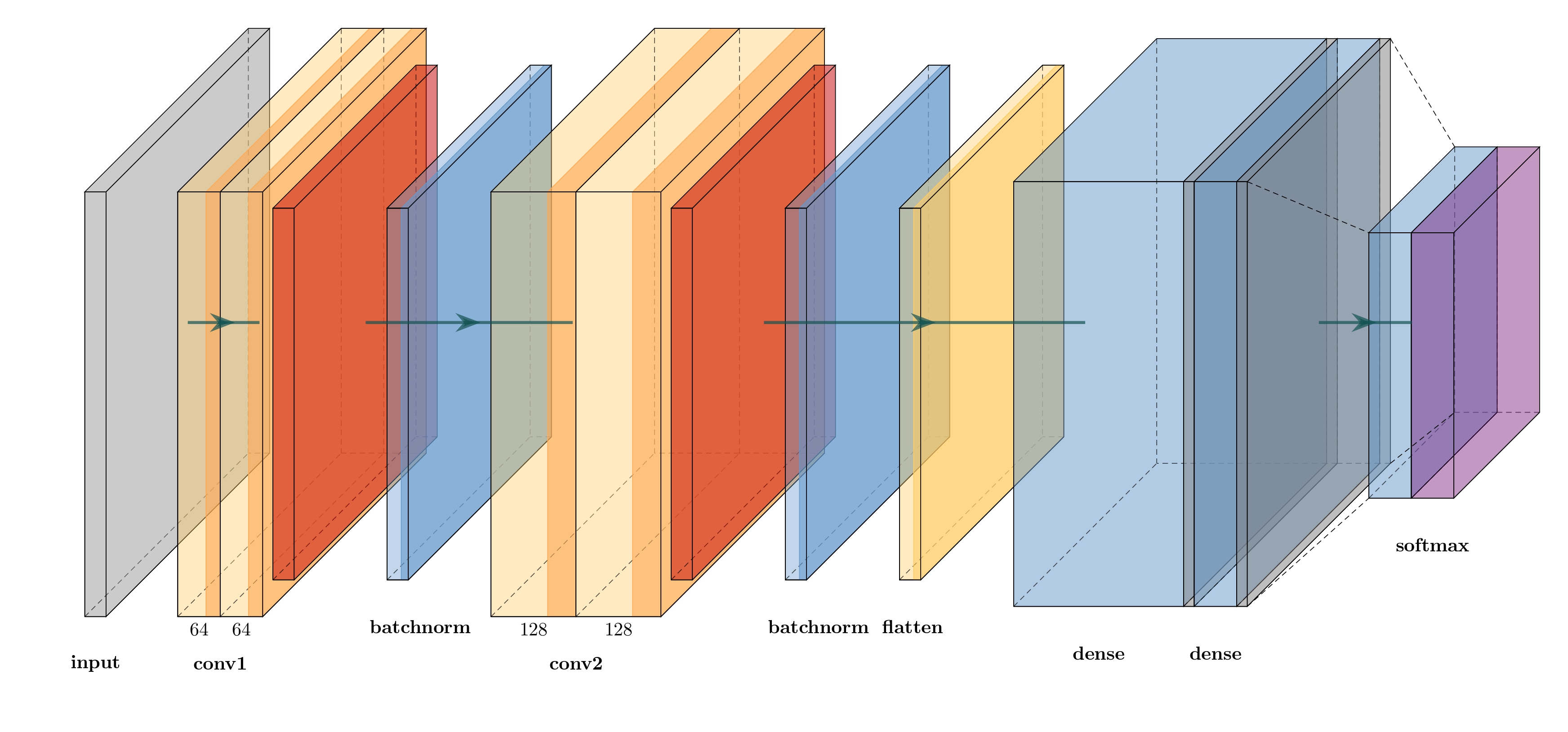}}
  \caption{Custom CNN architecture}\label{fig:cnn_architecture}
\end{figure}

The custom 2D convolutional classification model was built using the TensorFlow Keras API \cite{tensorflow,keras}. For this implementation, we ignore the z-dimension of the MediaPipe hand landmarks to reduce the complexity of the model. The model features two convolutional blocks, each consisting of two ReLU convolutional layers with a max pooling layer and a batch normalization layer. The output of the second convolutional block is flattened and fed into a set of dense and dropout layers prior to softmax activation. The model was trained using the Adam optimizer with a learning rate of 0.0005 and a sparse categorical cross-entropy loss function. It was trained for 100 epochs with a batch size of 128, achieving a training and validation accuracy of 99.67\% and 99.89\% respectively. \autoref{fig:cnn_architecture} illustrates the model architecture.

\subsubsection*{PointNet Architecture}

 PointNet \cite{PointNet} provides a point cloud classification architecture that is invariant to permutations, rotations, and translations of the input data. The model uses a series of multi-layer perceptrons (MLPs) to transform the input point cloud into a higher-dimensional feature space and then aggregates the features using max pooling layers. Then, it uses softmax activation to create a probability distribution over the output classes.

 \begin{figure}[!htbp]
  \centerline{\includegraphics[width=\linewidth]{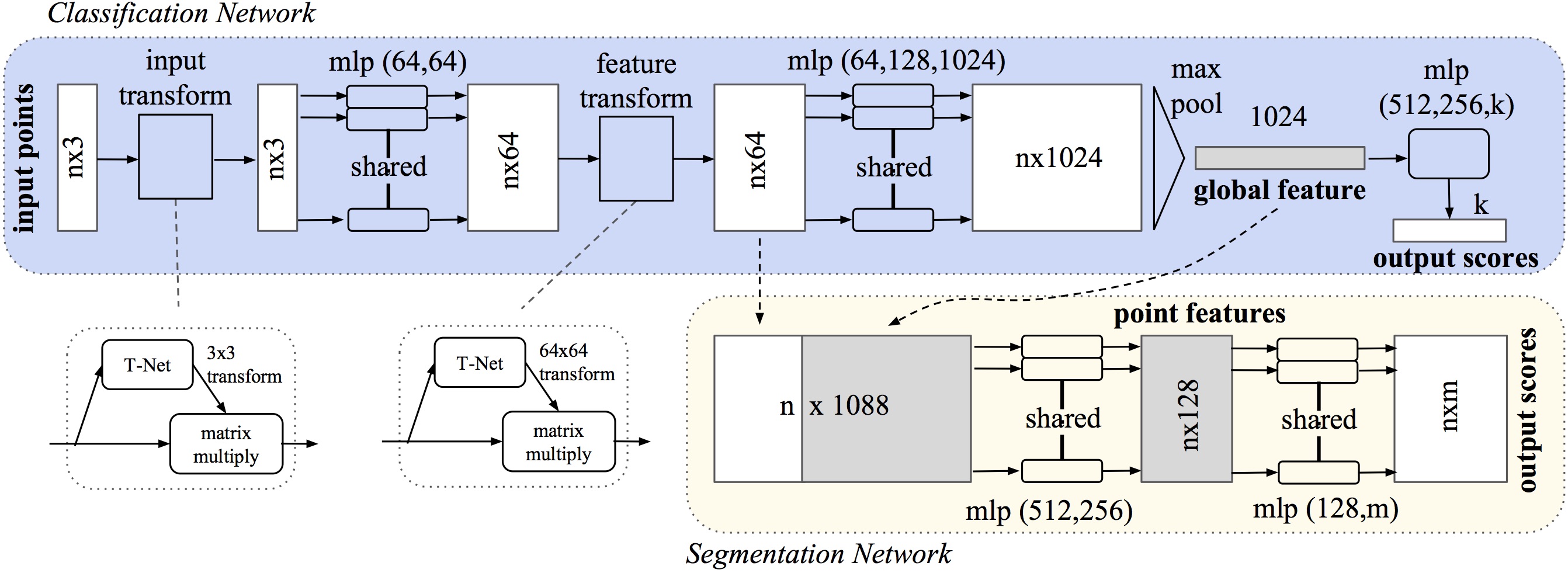}}
  \caption{PointNet Architecture}\label{fig:pointnet_architecture}
\end{figure}

 We implemented the 3D PointNet architecture using the TensorFlow Keras API \cite{tensorflow,keras}. The model was trained using the Adam optimizer with a learning rate of 0.0005 and a sparse categorical cross-entropy loss function. It was trained for 100 epochs with a batch size of 64, achieving a training and validation accuracy of 99.39\% and 99.43\% respectively. \autoref{fig:pointnet_architecture} illustrates the complete PointNet architecture.

\subsection{Model Inference}
\begin{figure}[!htbp]
  \centerline{\includegraphics[width=\linewidth]{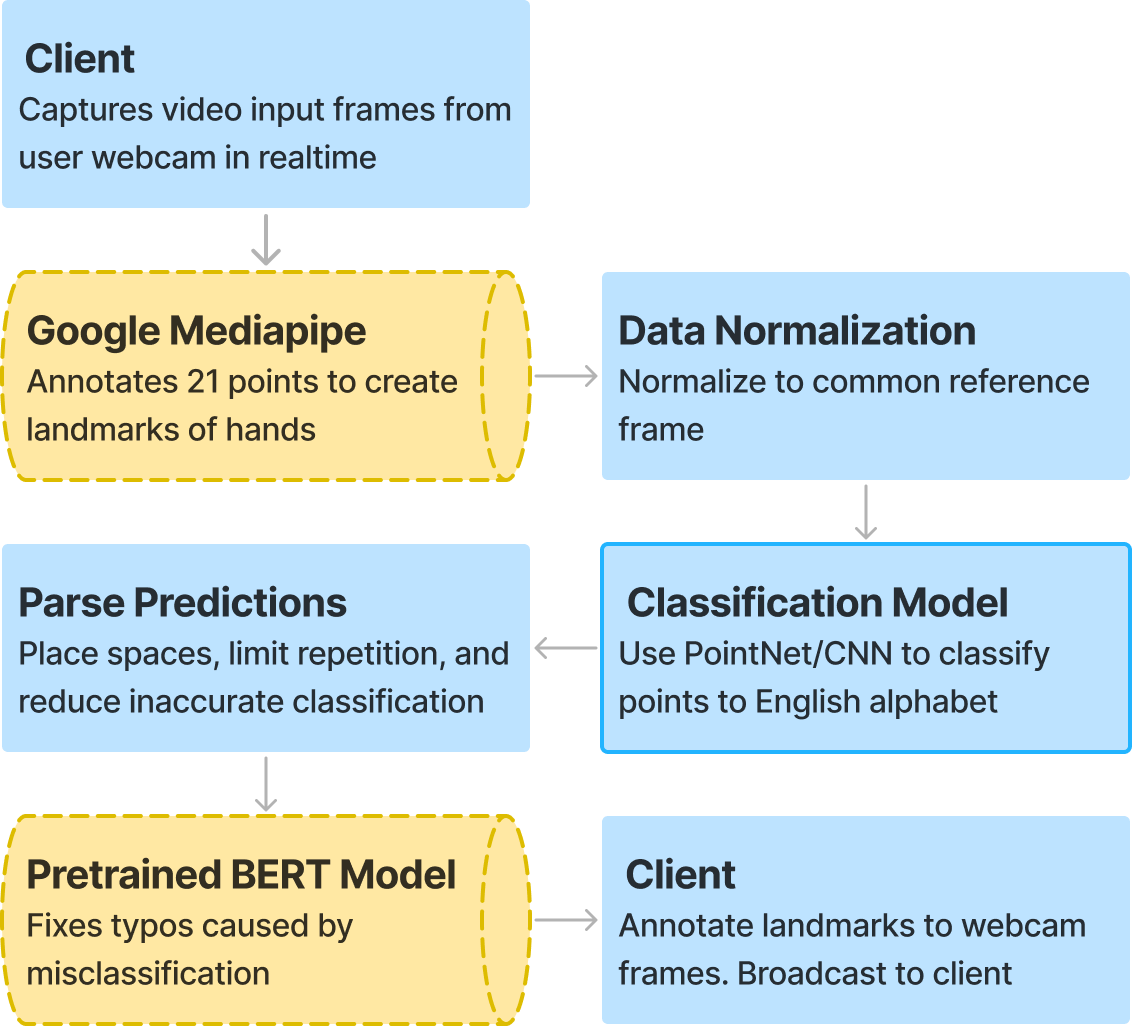}}
  \caption{Recognition infrastructure}\label{fig:recognition_flow}
\end{figure}

The inference stage of the model includes three main steps: fingerspelling-to-landmark conversion, landmark-to-text classification, and text synthesis. \autoref{fig:recognition_flow} demonstrates the complete inference architecture of the recognition component of the interface.

\subsubsection*{Fingerspelling \textrightarrow\ Landmark Conversion}
The Google MediaPipe Hand Landmark model is used to detect 21 landmarks in the inputted frames. The detected landmarks are then scaled and normalized to a common reference frame, ensuring consistency across different images, and making the model invariant to hand sizes and distances from the camera. Now, since the system is only dealing with the 21 landmarks, the computational resources required to run the model are significantly reduced, allowing for real-time fingerspelling recognition in real-world scenarios.
\subsubsection*{Landmark \textrightarrow\ Text Classification}
The normalized landmarks from the previous step are fed into either the 2D CNN model or the 3D PointNet model. The models classify the fingerspelling into individual letters, and the output is a sequence of English alphabets. We ensure that every fingerspelled letter is present over multiple consecutive frames to reduce the likelihood of misclassification. Furthermore, we use conditionals to differentiate between commonly misrecognized letters. For instance, since the letters $\mathbf{\{A, M, N, S, T\}}$ are commonly misrecognized, we use conditionals to distinguish the position of key landmarks like the thumb to correct these misclassifications.
\subsubsection*{Text Synthesis}
We insert a space between words when the hand is not present in the frame, and we prevent the repetition of the same letter more than twice consecutively. Finally, we use a pretrained BERT model to correct syntactical misclassifications and grammar errors. Overall, this process ensures that the model can accurately recognize fingerspelling and synthesize it into coherent spoken English text.

%% file: Production.tex
\section{Production}
\label{sec:production}

\subsection{Data Augmentation}
For the semantic gloss-to-pose retrieval step, we extracted skeletal poses from sign language videos using Google MediaPipe Holistic \cite{MediaPipe}. Furthermore, we used the all-MiniLM-L6-v2 embedding model \cite{minilm} to embed the glosses, and then stored the embeddings, glosses, and corresponding poses in a pgvector PostgreSQL database. This process allowed us to semantically retrieve and stitch together pose sequences for inputted gloss sequences.

\subsection{Inference}
\begin{figure}[!htbp]
  \centerline{\includegraphics[width=\linewidth]{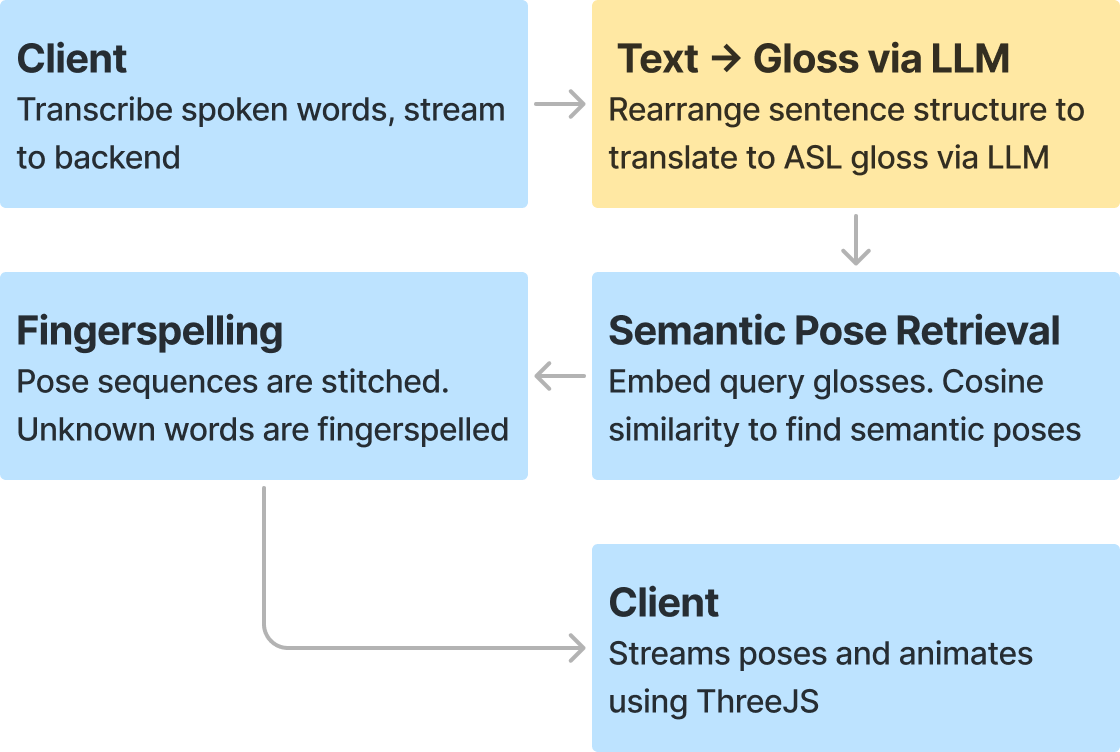}}
  \caption{Complete pose-retrieval infrastructure}\label{fig:production_flow}
\end{figure}

The inference stage of the sign language production component includes two main steps: text-to-gloss translation and semantic gloss-to-pose retrieval. \autoref{fig:production_flow} demonstrates the complete inference architecture of the pose retrieval component of the interface.

\subsubsection*{Text \textrightarrow\ Gloss Translation} We use a large language model (LLM) to translate spoken English text to ASL gloss. The LLM is instructed with a rule-based prompt which guides the model to generate ASL gloss by dropping articles and rearranging subjects, objects, and verbs. In the deployed interface, we use OpenAI's GPT-4o \cite{gpt4o} model because of its inference speed and accuracy. Alternative open-source LLMs can be easily integrated into the interface.
\subsubsection*{Semantic Gloss \textrightarrow\ Pose Retrieval}
The gloss sequences from the previous step are now individually embedded using the all-MiniLM-L6-v2 embedding model. These query vectors are used to semantically query the pgvector database with a certain cosine similarity threshold, allowing us to retrieve and stitch together contextually-similar pose sequences for inputted gloss sequences. For glosses that do not meet our cosine similarity threshold, we simply stitch together the poses of fingerspelling letters, allowing for the generation of poses for glosses not present in the data set.

%% file: Interface.tex
\section{Interface}
To demonstrate the model's capabilities in real-world settings, we present a demo interface for fingerspell recognition and semantic pose retrieval. The interface consists of two segments: a server and a client. The server provides a modular implementation of the recognition and production components that is accessible via a WebSocket API. The client is a web application that serves as a user-friendly interface for interacting with the two components.

\subsection{Server}
The server is implemented in Python using the Flask web framework. It provides a WebSocket API that allows clients to interact with the recognition and production components. The server is designed to be modular, allowing for easy integration with other existing applications. For receptive sign language inference, the server captures video streams, processes the frames using the recognition model, and emits the fingerspelled letters to the client through WebSocket messages. This architecture allows for real-time fingerspell recognition. For sign language production, the server receives spoken English text from the client, semantically retrieves and stitches the corresponding poses, and returns the corresponding ASL poses. This open-source API can be used to build a variety of applications, such as ASL translation systems, educational tools, and accessibility applications. See \autoref{sec:future_work} for potential applications.

\subsection{Client}
The client is a web application built with NextJS that provides a user-friendly interface for interacting with the server. It uses the WebSocket API provided by the server to communicate with the recognition and production components. For the fingerspell recognition component, the client displays a webcam stream which is annotated with recognized fingerspelling letters. It receives transcribed words from the server and displays it on the left half of the screen in real-time. For the pose production component, the client transcribes spoken English using the browser's built-in SpeechRecognition API\footnote{Audio transcription can be improved by using technology like OpenAI Whisper \cite{whisper}}, and sends the transcribed text to the server via WebSocket messages. The server then processes the text using the pose retrieval component and returns the corresponding ASL poses, which are animated on the right half of the screen using ThreeJS. The open-source \href{https://github.com/kevinjosethomas/sign-language-processing}{GitHub repository} features the full implementation and many examples of the demo interface.

%% file: FutureWork.tex
\section{Future Work}
\label{sec:future_work}

Sign language processing is a rapidly evolving field, and there is a growing interest in creating larger data sets and continuous sign language recognition systems that can handle the complexity and variability of natural sign language. This innovation will continue to be driven by advances in computer vision, machine learning, and natural language processing. In this section, we discuss potential future directions for expanding the current model and its applications.

\subsection*{Continuous Sign Language Recognition}
The current model is limited to recognizing individual fingerspelled letters. A natural extension is to recognize continuous ASL signs. This would require a more sophisticated model, such as an LSTM or a transformer model, that can capture the temporal dynamics of sign language. An interesting direction would be to explore the use of multimodal models that utilize parameter search to narrow down potential signs based on the hand placement, orientation, and other ASL parameters. By utilizing growing sign language corpora like ASL-Lex \cite{asllex}, models can be trained to identify parameters from poses and use them to reduce the search space and improve the accuracy of classification. Approaches could also project poses into an embedding space and use similarity metrics to classify signs. 

\subsection*{Natural Sign Language Production}
Another important direction is to improve the sign language production component's ability to express the more complex aspects of ASL, such as facial expressions, lexicalized signs, directional signs, and other non-manual markers. Additionally, unlike generating poses, an optimal model should be able to produce human-like signing motions that are fluid, natural, and expressive. One potential approach would be to use kinematic models to rig and control a virtual avatar's motions based on the output of the production module. Another approach would be to use generative diffusion models to produce photo-realistic humans signing based on the input text.

\subsection*{Consumer Applications}
While academia has made significant progress in sign language recognition and translation, there is a gap between research and real-world applications. As sign language processing technology matures, it is important to consider how it can be integrated into consumer applications and be deployed in the real-world. One potential implementation of the documented pose retrieval module is a Chrome extension that automatically embeds the sign language avatar in YouTube videos to provide ASL translations instead of relying on captions. Another example is a video-conferencing extension or webcam client that provides real-time ASL production for individuals who are speaking. While most research is still in the early stages of development and far from human parity, they already pose significant potential to improve accessibility.

%% file: Conclusion.tex
\section{Conclusion}
We have presented an open-source fingerspell recognition and semantic pose retrieval interface with the goal of advancing sign language translation systems. The interface combines convolutional neural networks and pose estimation models to create a system for translating between ASL fingerspelling and spoken English. The fingerspell recognition module translates ASL fingerspelling into spoken English, while the semantic pose retrieval module converts spoken English into ASL poses. The interface's ability to function reliably in real-time and under diverse environmental conditions, such as varying skin tones, backgrounds, and hand sizes, is an important step towards making sign language translation more accessible and inclusive. This adaptability ensures that the technology can be effectively used by a broader audience without the need for specialized environments or equipment, which is crucial for real-world applications.

Moreover, the modular design of the interface allows for easy integration into various platforms, making it a versatile tool for enhancing accessibility and inclusivity in a wide range of existing applications. We hope that this project will invite further collaboration and innovation in the field of sign language translation, inspiring developers to build upon the existing system and create more advanced solutions that capture the richness and complexity of sign language communication. By engaging with the Deaf community and incorporating their feedback, we can continue to improve the interface and ensure that it meets the diverse needs of its users.

While the current system provides a solid foundation for sign language translation, it more importantly serves as a stepping stone towards more advanced ASL translation systems. Fingerspelling is only one small aspect of ASL, and there is much more work to be done to capture the full richness and complexity of sign language grammar and syntax. By building upon this foundation, we can create more sophisticated systems that capture the nuances of sign language communication and provide a more inclusive and accessible experience for all individuals.